# Comparing PSDNet, pretrained networks, and traditional feature extraction for predicting the particle size distribution of granular materials from photographs

**Javad Manashti**[1], **François Duhaime**[2], **Matthew F. Toews** [3], **Pouyan Pirnia**[4], **Jn Kinsonn Telcy**[5]

Abstract

This study aims to evaluate PSDNet, a series of convolutional neural networks (ConvNets) trained with photographs to predict the particle size distribution of granular materials. Nine traditional feature extraction methods and 15 pretrained ConvNets were also evaluated and compared. A dataset including 9600 photographs of 15 different granular materials was used. The influence of image size and color band was verified by using six image sizes between 32 and 160 pixels, and both grayscale and color images as PSDNet inputs. In addition to random training, validation, and testing datasets, a material removal method was also used to evaluate the performances of each image analysis method. With this method, each material was successively removed from the training and validation datasets and used as the testing dataset. Results show that a combination of all PSDNet color and grayscale features can lead to a root mean square error (RMSE) on the percentages passing as low as 1.8 % with a random testing dataset and 9.1% with the material removal method. For the random datasets, a combination of all traditional features, and the features extracted from InceptionResNetV2 led to RMSE on the percentages passing of 2.3 and 1.7 %, respectively.

1 Laboratory for Geotechnical and Geoenvironmental Engineering (LG2), École de technologie supérieure, Montréal, Quebec, H3C 1K3, Canada; https://orcid.org/0000-0001-9879-3931
2 Laboratory for Geotechnical and Geoenvironmental Engineering (LG2), École de technologie supérieure, Montréal, Quebec, H3C 1K3, Canada; https://orcid.org/0000-0001-8289-5837
3 Systems Engineering Department, École de technologie supérieure, Montréal, Quebec, H3C 1K3, Canada; https://orcid.org/0000-0002-7567-4283
4 Laboratory for Geotechnical and Geoenvironmental Engineering (LG2), École de technologie supérieure, Montréal, Quebec, H3C 1K3, Canada; https://orcid.org/0000-0003-2939-2813
5 Laboratory for Geotechnical and Geoenvironmental Engineering (LG2), École de technologie supérieure, Montréal, Quebec, H3C 1K3, Canada

Introduction

In geotechnical engineering, the particle size distribution (PSD) of particles larger than 0.075 mm is usually determined through sieving, according to ASTM D6913, *Standard Test Methods for Particle-Size Distribution (Gradation) of Soils Using Sieve Analysis*. Sieving is one of the most common tests conducted in commercial laboratories. However, sieving is time-consuming and energy-intensive. It is also comparatively expensive with regards to equipment maintenance and replacement of defective or worn sieves. The volume of water required for washing is also high (Ohm 2013). Conversely, image-based methods are eco-friendly, sustainable, and quick, with no need for water and low energy consumption. The benefits of PSD determination by image analysis over sieving include improved laboratory environment, reduced energy consumption, and shorter test time (Ohm 2013).

Image analysis methods for PSD determination can be divided in two main categories: direct and indirect methods. Direct methods use image segmentation. The contour of each particle in the photograph is traced. Statistical relationships are used to relate the size distribution of the segmented areas with the PSD. Commercial codes such as WipFrag (Maerz, Palangio, and Franklin 1996) are often centered on direct methods. Indirect methods use features that describe image texture to estimate the PSD. Textural features describe the spatial structure of pixel intensity (Liu et al. 2019; Tuceryan and Jain 1993).

Textural features can be classified into three main categories (Tian 2013). The first category uses statistical parameters to describe the spatial distribution of pixel intensity. For example, Haralick features (Haralick, Shanmugam, and Dinstein 1973) use statistics (e.g., average, variance, correlation) of a global co-occurrence matrix that defines the number of pixel pairs separated by a specific offset for each combination of pixel intensity. Features of the second category are based on local patterns found in the image. One example is the Local Binary Pattern (LBP) method (Ojala, Pietikainen, and Maenpaa 2002). With this approach, the relative intensity of pixels in the neighborhood of a center pixel are encoded as a binary integer. Transform-based methods form the third category. Gabor wavelets (Manjunathi and Ma 1996) and Fourier transforms (Szeliski 2011; Yaghoobi et al. 2019) are included in this category. Transformation-based methods allow the extraction of information on the distribution of pixel intensity in the frequency domain (Shin and Hryciw 2004).

Several studies have appraised the performances of PSD determination methods based on textural features. Using a dataset of 53,003 synthetic images of granular material prepared with the discrete element code YADE ( Pirnia et al. 2019), Manashti et al. (2021) reviewed the performances of nine textural parameters from the three main categories. The textural features were used as inputs for a series of artificial neural networks. All nine methods could perform relatively well with idealized spherical particles, pending an optimization of their parameters. A selection of the 618 best features from the three main categories gave a root mean square error (RMSE) on the percentages passing of 3.4 % and a relative error on the $D_{50}$ of 6.1 %, where $D_{50}$ is the particle size for which 50 % of the sample mass is composed of smaller particles. Performances with real photographs can also be estimated from published data. Ohm and Hryciw (2014a) used a semi-empirical relationship based on Harr wavelet transforms to estimate the PSD of granular material from photographs after mechanical sorting in a sedimentation column. An RMSE of 7.4 % for the percentages passing and a relative error of 7.6 % for the $D_{50}$ can be estimated from their results.

Convolutional neural networks (ConvNets) provide a new artificial neural network structure for inputs consisting of images and sounds (Abdel-Hamid et al. 2012; Chellapilla, Puri, and Simard 2006; Kavukcuoglu et al. 2010; Krizhevsky, Sutskever, and Hinton 2012; LeCun and Bengio 1995). ConvNets apply a series of filters on the images to extract deep data that is fed to a fully connected neural network. A typical ConvNet uses a large number of layers with millions of weight parameters and thousands of training images and tuning parameters (Lee et al. 2017). Contrarily to the method used by Manashti et al. (2021) with traditional textural features fed to a fully connected network, ConvNets do not use engineered features but automatically determines the best features from the images to minimize the error rate (Fu and Aldrich 2018). ConvNet are used in many areas, including medical imaging (Hall et al. 2010; Lee et al. 2017; Plautz et al. 2017; Tajbakhsh et al. 2016; Yan, Lu, and Summers 2018), speech recognition (Abdel-Hamid, Deng, and Yu 2013; Abdel-Hamid et al. 2012; LeCun and Bengio 1995), and soil classification and geochemistry (Liu, Ji, and Buchroithner 2018; Ng, Minasny, and McBratney 2020; Padarian, Minasny, and McBratney 2019). The most well-known ConvNet models were developed with the ImageNet dataset (Deng et al. 2009) to identify common objects such as computers, cups, and animals. The performances of ConvNets are generally superior to conventional approaches based on engineered features (Fu and Aldrich 2018).

Pirnia, Duhaime, and Manashti (2018) made the first attempt at using ConvNets to predict the PSD of granular materials based on images in geotechnical engineering. They used the same synthetic image dataset as Manashti et al. (2022). Grayscale images of 128×256 pixels were used. The model structure was composed of four ConvNet blocks with 5×5 filters followed by a rectified linear unit (ReLU) layer. Each pair of convolutional blocks was followed by a max pooling layer. The four convolutional blocks were followed by three fully connected layers. The RMSE for all sieves, the finest, and coarsest sieve were 6.9, 4.2, and 9.1 %, respectively. The ConvNet model presented by Pirnia, Duhaime, and Manashti (2018) was later improved and forms the basis of PSDNet, the model presented in this paper. Compared to Pirnia, Duhaime, and Manashti (2018), PSDNet was improved by increasing the number of neurons in the fully connected layers, by using max pooling layer after each convolutional block, by adding dropout layers in the convolutional blocks and by increasing the number of filters in the convolutional blocks. The influence of image size (32 to 160 pixels), color band (color or greyscale image) and viewpoints (photographs taken from the top or through the bottom of a transparent container) was studied. For the color dataset, the RMSE on the percentages passing for all sieves, the coarsest sieve and the finest sieve were 2.8, 3.3, and 1.8 %, respectively. The relative error on $D_{50}$ was 5.6 %. It should be noted that poorer performances for large particles were also observed by Manashti et al. (2021) with traditional features.

Buscombe (2020) introduced SediNet, a ConvNet model aimed at classifying sediment images and determining their PSD. SediNet includes four convolutional blocks, each consisting of between 16 and 64 convolutional filters, a batch normalization layer, and a max pooling layer. A regression model was trained to predict nine grain size percentiles including the $D_{50}$. The RMSE on the percentiles varied between 24 and 45 %. The predictions of SediNet were also compared to those of direct image analysis using image segmentation and texture analysis using wavelets for a small dataset of beach sand photographs by McFall et al. (2020). The mean error on $D_{50}$ for SediNet was 22 % compared to 34 and 36 % for image segmentation and wavelet analysis.

GRAINet is another ConvNet model that was developed to predict the PSD of sediments (Lang et al. 2021). It was trained with images featuring a grain size range of 0.5 to 40 cm from 25 gravel bars along six rivers in Switzerland. GRAINet has an entry ConvNet layer of 3×3 followed by six convolutional blocks and a ConvNet layer of 1×1. The model was used to predict the PSD and the mean $d_m$ values of the gravel bars, where $d_m = \sum P_i d_i /100$ with $d_i$ and $P_i$ as the mean size and percentage passing for bin $i$ of the PSD. An RMSE of 27 % was obtained for $d_m$.

Both GRAINet and SediNet used a manual segmentation of each image in their dataset to determine the ground truth PSD. SediNet was trained with a dataset including 409 images of 512×512 pixels. Lang et al. (2021) trained GRAINet with a dataset including 1,491 images augmented with vertical and horizontal flipping. McFall et al. (2020) used 63 images in their comparison of SediNet with classical direct and indirect methods. The number of images in their dataset was increased to 517 by flipping the images horizontally and by using a moving window to sample different parts of each image. The datasets presented by these authors are small compared to those used to train networks for object classification. These larger datasets often include several hundred thousand images. As observed by Lang et al. (2021), networks trained with small datasets can suffer from overfitting.

Synthetic images offer one avenue to increase the size of image datasets. The dataset presented by Pirnia, Duhaime, and Manashti (2018) was more than one order of magnitude larger than the current real photographs datasets. However, it comprises highly idealized images of spherical particles. This type of dataset allows the influence of important parameters to be verified, for example the influence on RMSE of using both views from the top and bottom of a transparent container. However, the relationship between the performances obtained for synthetic images and real images remains poorly characterized (Duhaime et al. 2021).

Pretrained networks offer another approach to train ConvNets for applications where gathering large datasets is difficult. With pretrained networks, a ConvNet trained on a different dataset for a different task, often object classification with the ImageNet dataset, is applied to a different task in a different domain (Fu and Aldrich 2018). Two main methods can be used. With the first method, the network trained on a different task acts as a feature extractor. With this approach, the output of one of the fully connected layers can be used as a set of features (e.g. Fu and Aldrich 2018). With the transfer learning method, the new dataset is used to update a subset of the weights or all the weights in the pretrained model (e.g., Fu and Aldrich 2018; Peng et al. 2019). The extent of network training with the new dataset depends on the similarity of the two tasks. For similar tasks, only the last layers need to be retrained as they are associated with higher level tasks (Tajbakhsh et al. 2016). The first layers are associated with lower level tasks akin to the determination of textural features with classical methods.

Manashti et al. (2022) used pretrained ConvNets to predict the PSD from synthetic images of granular materials. They evaluated different pretrained models with transfer learning and feature extraction methods. The best results for transfer learning models were obtained with densenet201 with RMSE values on the percentages passing of 2.3, 4.1, and 3.6 % for fine, coarse, and all mesh sizes, respectively. For feature extraction, InceptionResNetV2 reached RMSE values of 2.5, 4.4, and 3.4 % on the percentages passing for fine, coarse, and all mesh sizes.

This study aims at comparing the performances of various image analysis methods to predict the PSD of granular materials from photographs. The tested methods include ConvNet model PSDNet, 15 different pretrained ConvNets with feature extraction and transfer learning methods, and nine traditional feature extractors. A dataset comprising 9600 photographs of 15 different granular materials is introduced in this article. A material removal method was used to compare the performances of each method when confronted with a new material unseen during network training. Most of the methods presented in this paper have never been applied to a real photograph dataset in the context of geotechnical engineering.

## Methodology

### IMAGE DATASET

A dataset containing thousands of images is essential to train ConvNet models like PSDNet. Three materials were selected to obtain the dataset for this project: commercial Bomix sand, 2C sand and till. The 2C sand and till were described by Dumberry, Duhaime, and Ethier (2018). A total of five materials were obtained by sieving the Bomix and 2C sands: Bomix sand retained on the 425 μm sieve, Bomix sand passing the 425 μm sieve, 2C sand retained on the 625 μm sieve, 2C sand passing the 625 μm sieve, and till. These five materials and all possible 1:1 combinations were used to generate 15 granular materials with different PSD (Figure 1). The PSD were obtained using sieves of 75, 150, 300, 425, 600, 630, 850, 1185, 2360, and 4750 μm.

Each material was poured in a square pan and photographed 20 times to obtain a total of 300 images for the 15 materials. The materials were mixed thoroughly in the pan between each photograph. The photographs were taken using a Nikon D90 camera with a resolution of 2848 × 4288 pixels. A nominal focal distance of 55 mm and a sensitivity of ISO 400 were used. More details on the preparation of the dataset were provided by Duhaime et al. (2021).

To increase the size of the dataset, each image was divided into 32 subimages (4 rows × 8

columns) of 512×512 pixels to create 9600 images. The 512×512 images have a scale of 56 microns per pixel. This image sampling adds some errors on the PSD. Because of segregation during mixing, each sampled region should correspond to a slightly different PSD (e.g. Dubé et al. 2021). Moreover, the subimages situated in the corners of the primary images can be slightly out of focus and blurry (e.g. photograph 1 in Figure 2). These sampling errors are assumed small compared to the errors associated with the PSD determination with the image analysis techniques. Figure 2 shows some examples of images for the 15 materials.

**Figure 1** Particle size distribution of the 15 granular materials.

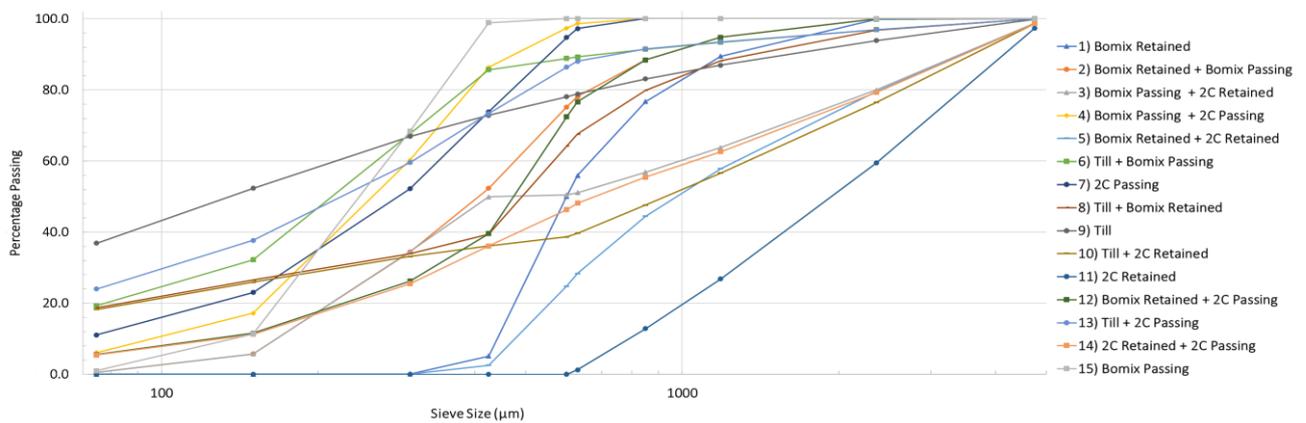

**Figure 2** Image samples for the 15 granular materials (512×512 pixels, the numbers correspond to those presented in Figure 1).

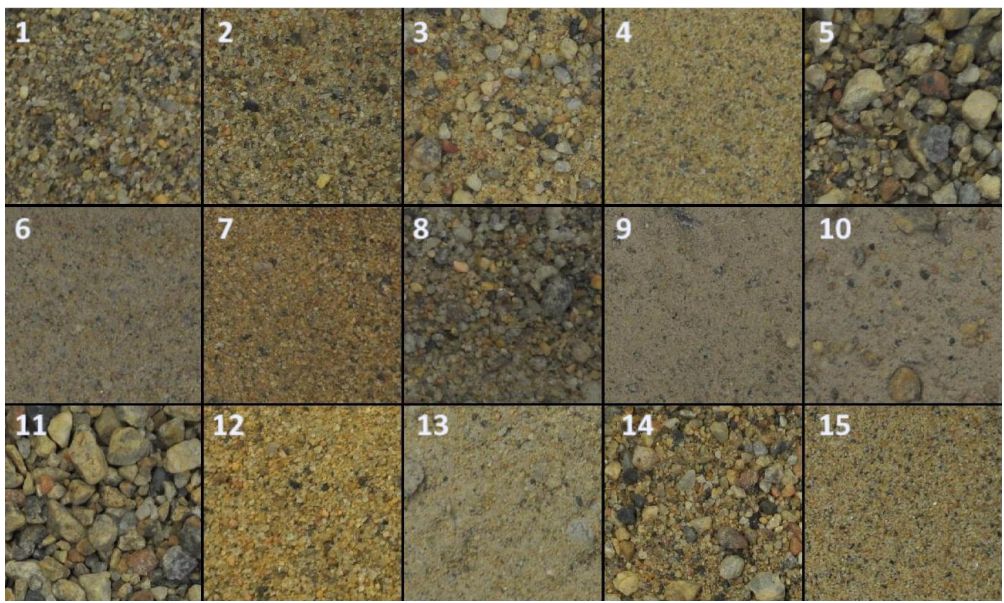

**PSDNET**

PSDNet was developed by Manashti et al. (2022) to predict the PSD of granular materials from images. It was developed using a large dataset including 53 103 synthetic images obtained with the discrete element method. In this paper, the PSDNet networks developed with the synthetic image dataset were trained from scratch to predict the PSD from real photographs of granular materials.

PSDNet was also implemented as a feature extractor. The last fully connected layer before the regression layer provided 100 features for each image. These features allowed the training of a series of smaller and more computationally efficient artificial neural networks (ANN). Usage of pretrained convolutional networks as feature extractors and ANN training is introduced in more detail later in the paper.

Table 1 gives the structure of PSDNet for color images of 160×160 pixels. PSDNet is centered on four convolutional blocks, each consisting of one convolutional layer and three fully connected layers. The three fully connected layers include batch normalization, ReLU, and max pooling layers. A dropout layer is also added after blocks 2 and 4. The four convolutional blocks are followed by three fully connected blocks. The fully connected blocks each include fully connected, batch normalization, ReLU, and dropout layers. The last block includes a fully connected layer and regression output to predict the percentages passing for the 10 sieves used for the PSD determination. PSDNet was described in detail by Manashti et al. (2022), together with the fine-tuning process for different image sizes. In the present study, PSDNet is used without any change to train and predict grayscale and color images of 32, 64, 96, 128, and 160 pixels. Only the top view was captured and used.

**Table 1** PSDNet layers for inputs consisting pf RGB images of 160×160 pixels.

| PSDNET LAYERS | DESCRIPTION |
| --- | --- |
| IMAGE INPUT | 160×160×3 images with 'zero center' normalization |
| CONVOLUTION | 160 5×5×3 convolutions with stride [1 1] and padding [1 1] |
| BATCH NORMALIZATION | Batch normalization with 160 channels |
| RELU | ReLU |
| MAX POOLING | 3×3 max pooling with stride [2 2] and padding [0 0 0 0] |
| CONVOLUTION | 320 5×5×160 convolutions with stride [1 1] and padding [1 1] |
| BATCH NORMALIZATION | Batch normalization with 320 channels |
| RELU | ReLU |
| MAX POOLING | 3×3 max pooling with stride [2 2] and padding [0 0 0 0] |
| DROPOUT | 20% dropout |
| CONVOLUTION | 640 5×5×320 convolutions with stride [1 1] and padding [1 1] |
| BATCH NORMALIZATION | Batch normalization with 640 channels |
| RELU | ReLU |
| MAX POOLING | 3×3 max pooling with stride [2 2] and padding [0 0 0 0] |
| CONVOLUTION | 640 5×5×640 convolutions with stride [1 1] and padding [1 1] |
| BATCH NORMALIZATION | Batch normalization with 640 channels |
| RELU | ReLU |
| MAX POOLING | 3×3 max pooling with stride [2 2] and padding [0 0 0 0] |
| DROPOUT | 20% dropout |
| FULLY CONNECTED | Fully connected layer with 1500 nodes |
| BATCH NORMALIZATION | Batch normalization with 1500 channels |
| RELU | ReLU |
| DROPOUT | 20% dropout |
| FULLY CONNECTED | Fully connected layer with 500 nodes |
| BATCH NORMALIZATION | Batch normalization with 500 channels |
| RELU | ReLU |
| DROPOUT | 20% dropout |
| FULLY CONNECTED | Fully connected layer with 100 nodes |
| BATCH NORMALIZATION | Batch normalization with 100 channels |
| RELU | ReLU |
| DROPOUT | 20% dropout |
| FULLY CONNECTED | Fully connected layer with 10 nodes |
| REGRESSION OUTPUT | Mean square error with response' Response' |

**TRADITIONAL FEATURE EXTRACTION**

In this study, nine traditional feature extraction methods were used to generate features that were fed to an ANN. The MATLAB codes developed by Manashti et al. (2021) were used after minor modifications. Grayscale images with a resolution of 512 × 512 pixels were used to train and predict for traditional feature extraction.

**Histogram of Oriented Gradients (HOG)**

Histograms of oriented gradients (HOG) describe an image with a count of the different orientations of the gradient of gray level intensity for a series of predefined square or rectangular cells (Dalal and Triggs 2005). For each cell, the gradients are allocated to a series of bins. The gradient magnitude is used to weight the gradient when it is assigned to the bins. Larger blocks with several cells are also defined to normalize the content of each bin according to the norm of all bins in the block. Normalization allows for the consideration of local contrast variations in an image. Cells that are smaller or of a similar size to the particles are usually dominated by a small number of particle edges that give the gradient a preferential orientation. More random gradient orientations will result from cells that are larger than the particles (Pang et al. 2011).

In the present study, HOG were calculated for nine orientation bins and cells of 2, 4, 8, 16, 32, and 64 pixels. The MATLAB function *extractHOGFeatures* was used. The mean and standard deviation was calculated for all HOGs to generate 12 features for the ANN inputs. The histograms were normalized with blocks of four cells with an overlap of a cell width for contiguous blocks.

**Local Entropy of the grayscale image**

Entropy is a statistical parameter that measures the randomness of the gray-level intensity distribution in the neighborhood of a pixel (Gonzalez and Woods 2002). Lower local entropy values, or more constant gray levels, are expected to be associated with large particles. The nine filters presented by Manashti et al. (2021) were used to define the neighborhood around the center pixel. The

MATLAB functions *Getnhood* and *Strel* were used to set these filters. Entropy was calculated with the *entropyfilt* function. Entropy features were calculated for bin sizes of 4, 8, 16, 32, 64, 128, and 256 pixels. The mean and standard deviation of the entropy for the nine disk filters and the seven bin sizes were computed and applied as the ANN inputs (total of 126 features).

**Local Binary Pattern (LBP)**

With local binary patterns (LBP), the pixels of a neighborhood are compared to a threshold set by the centre pixel (Ojala, Pietikainen, and Maenpaa 2002). The pixels in the neighborhood take values of 0 or 1 depending if their gray level is higher or lower than the threshold value (Sudha and Ramakrishna 2017). This pattern is coded in a binary number of $p$ bits, where $p$ is the number of pixels in the neighborhood around the center pixel. Different numbers of neighboring pixels and distances between the center and the neighboring pixels may be selected (Kaya, Ertuğrul, and Tekin 2015). The binary numbers that only differ by a rotation can be combined into bins to reach rotation invariance, and thus limit the number of features. The MATLAB functions *extractLBPFeatures* was utilized to calculate LBP. In this study, LBP was computed for 20 neighbors at a radius of 3 pixels for a total of 383 features. These LBP features were used as ANN inputs.

**Local configuration pattern (LCP)**

Local Configuration Patterns (LCP) combine LBP with Microscopic configuration of images (MiC) (Deng and Yu 2015; Guo, Zhao, and Pietikäinen 2011). LBP does not describe the variability of the gray level intensity. MiC provides this information by finding the linear combination of neighboring pixel intensity that best describes the intensity of the center pixel. This linear combination is defined for each rotation invariant LBP bin. A least square approach is used to calculate the weights of the linear combinations. The LCP features that are fed to the ANN are a combination of the LBP histogram and MiC. In this study, the LCP features were calculated for 20 neighboring pixels at a radius of 3 pixels, for a total of 441 features. The calculations were made with the MATLAB script described by Manashti et al. (2021). This scripts uses the original LCP code developed by Guo, Zhao, and Pietikäinen (2011).

**Completed Local Binary Pattern (CLBP)**

The Completed Local Binary Pattern (CLBP) has three parts (Guo, Zhang, and Zhang 2010). The first part corresponds to the LBP. The second part gives information on the absolute value of the gray level differences between the center pixel and the neighboring pixels. A binary number is obtained from the differences by thresholding them with the mean absolute difference for the whole image. The third part describes the center pixel intensity. It is thresholded with the mean pixel intensity to obtain another binary value. The binary numbers are then combined and used to build a histogram. The CLBP was computed for 20 neighbors at a radius of 3 pixels, similarly to LBP and LCP. The total number of features for CLBP is 44. The MATLAB script presented by Manashti et al. (2021) was used to calculate the CLBP features. It is centred on the original CLBP code of Guo, Zhao, and Pietikäinen (2012).

**Fourier Transform**

Image features can be calculated based on the mean and standard deviation of the magnitude spectrum of 2D Fourier transforms for a series of ring filters (Szeliski 2011; Yaghoobi et al. 2019). Ten rings filters were used. The thickness of successive rings was increased by a factor 1.6. More details on the Fourier transform and ring filters are presented by Manashti et al. (2021). The mean and standard deviation of the magnitude spectrum of the Fourier transform allowed 20 features to be defined for the ANN inputs.

**Gabor filters**

Gabor filters with different wavelengths and orientations were applied to the images. Gabor filters combine the Gaussian kernel function with a sinusoid wave (Tuceryan and Jain 1993; Yaghoobi et al. 2019), enabling the local frequency content of an image to be evaluated. The MATLAB Gabor function was used for the preparation of a filter bank corresponding to wavelengths of 3 to 15 pixels, and orientations between 0 and 150°. The mean and standard deviation of 42 filters (84 parameters) were used as ANN inputs. The MATLAB function *imgaborfilt* was used to apply the filter bank to the images.

**Haar wavelet transforms**

Hryciw et al. (2015) presented detailed calculations of two-dimensional Haar wavelet transforms in the context of PSD determination. Each level of the wavelet transform reduces the image size by a factor of 2. For each level and for each 4×4 pixel neighborhood in the image, four coefficients are calculated using the mean pixel intensity ($A$), and pixel intensity differences between rows ($H$), columns ($V$) and diagonals ($D$). The sum of the square of the $A$, $H$, $V$, and $D$ coefficients for the whole image describes the image energy. The total energy is preserved for each level of Haar wavelet transform for a given image (Hryciw et al. 2015). To define a series of features, the sum and standard deviation of $H^2$, $V^2$, and $D^2$ were calculated for 7 levels of wavelet transform. This results in 42 features. The coefficients were calculated using a MATLAB code developed by Manashti et al. (2021) and centred on the MATLAB function *haart2*.

**PRETRAINED CONVNET**

Due to the large number of weights in their structure, ConvNets require more resources to train compared with other types of neural networks. When trained from scratch, the initial weights are usually generated using random numbers with a normal distribution, a mean of 0 and a small standard deviation. In these circumstances, pre-trained ConvNet models that have been trained on similar datasets, or even very different datasets in our case, can be adapted to save time and computing resources.

In this article, 15 ConvNet models were evaluated for transfer learning and feature extraction. The tested models include AlexNet (Krizhevsky, Sutskever, and Hinton 2012), VGG16 (Simonyan and Zisserman 2014), VGG19 (Simonyan and Zisserman 2014), GoogLeNet (Szegedy et al. 2015), InceptionV3 (Szegedy et al. 2016b), DenseNet201 (Huang et al. 2017), MobileNetV2 (Sandler et al. 2018), ResNet18 (Wu, Zhong, and Liu 2018), ResNet50 (He et al. 2016), ResNet101 (He et al. 2016), Xception (Chollet 2016), InceptionResNetV2 (Ioffe and Szegedy 2015; Szegedy et al. 2016a), ShuffleNet (Zhang et al. 2017), NasNetMobile (Zoph et al. 2018), and NasNetLarge (Zoph et al. 2018). Depending on the structure of the model, four different sizes of images (224, 227, 299, and 331) were used for training and validation of pretrained convnets. Three methods were used for each model.

With the first method, the final SoftMax (classification layer) of the 15 networks was replaced by a regression layer to predict the percentages passing for the 10 sieve sizes. Starting from the pretrained weights, each model was trained for five epochs. Because of the fundamental difference between the tasks considered in this paper (PSD determination) and the original task with the ImageNet dataset (object classification), the weights were updated for all layers. This first method will be referred to as "transfer learning".

With the second method, the pretrained models that were retrained using the material photograph dataset with the first method were used as feature extractors. A total of 1,000 features were extracted from the last fully connected layer. The PSD were predicted using an ANN created with the *fitnet* function in MATLAB to calculate the percentages passing for the 10 sieve sizes. The second method will be referred to as "feature extraction of transfer learning".

With the third method, the last fully connected layer of each pretrained model was used to extract 1000 features without retraining the pretrained ConvNets with the granular material photographs. The main difference between the second and third methods is the absence of ConvNet training with the material photographs for the third method. As with the second method, the features were fed into an ANN created with the *fitnet* function in MATLAB. This method will be referred to as "feature extraction".

It is expected that feature extraction of transfer learning will lead to better results than pure feature extraction because of the ConvNet training. The feature extraction of transfer learning allows for a minimal retraining of the ConvNet layers that accomplish lower level tasks, and a more extensive training of the fully connected layers in the ANN that execute higher level tasks. The ANN training with feature extraction methods is much faster than the ConvNet training.

## TRAINING PARAMETERS

The training of PSDNet, the pretrained models, and the feature extraction ANN was conducted using MATLAB 2020b. For each method, the network performances were calculated based on the RMSE of the percentages passing for all sieves. The following sections give the training parameters for each network.

### PSDNet

The same parameters as Manashti et al. (2022) were used to train PSDNet. The initial learning rate was set to 0.0001 and the number of epochs was set to 60. Weights were updated with the Stochastic Gradient Descent with Momentum (SGDM). The training data was shuffled after every epoch during training. Predictions were made for 10 sieve sizes (75, 150, 300, 425, 600, 630, 850, 1185, 2360, 4750 µm). Training was conducted using two graphical cards (Nvidia GeForce RTX 2080 and GTX 1650). Compared with Manashti et al. (2022), one significant difference is the shorter time needed to train PSDNet because of the lower number of images (9600 rather than 53003).

Network performances are determined using a test dataset. This subdataset it not shown to the neural network during training and validation. Two types of test datasets were used. With the first method, the test dataset comprises 10 % of the images and is randomly sampled. This will be referred to as the random testing dataset in the paper. The 15 materials are included in the training, testing, and validation datasets with this approach. 7680 photographs (80%) were randomly selected as the training dataset, 960 images (10%) as validation dataset, and the same number as the test dataset. The second method will be referred to as the material removal test. It was designed to verify the performances of ANN and ConvNets if images of a new material, unseen during training, are fed to the networks. With this method, the 15 granular materials were successively removed from the training and validation datasets and used as the testing dataset. The ability of each method to predict the PSD of unseen materials is expected to give a more realistic portrait of their performances. With the material removal test, the test dataset contains 640 images of a given material. The 14 remaining materials were sampled randomly to fill the training (8320 images) and validation datasets (640 images). The material removal technique was inspired by the geographical cross-validation used by Lang et al. (2021).

**Pretrained models**

Pretrained models followed PSDNet parameters and specifications for the training, validation and test process. However, in pretrained models, five training epochs were performed, and the initial learning rate of 0.000001 was applied. The size of the images varied according to the pretrained model (224, 227, 299, and 331 pixels). Training, validation and testing was conducted using both random testing datasets and the material removal test.

**ANN for traditional features and feature extraction**

Function fitting neural networks (*fitnet*) were used in MATLAB 2020b to predict the percentage passing for the 10 sieves using extracted features. The hyperbolic tangent sigmoid transfer function was used as with the ANN models presented by Manashti et al. (2021). Network training was conducted with the scaled conjugate gradient method. The maximum number of training epochs was set to 10,000 for all networks.

The numbers of layers, and neurons on each layer were determined by trial and error. They vary with the number of inputs, which ranged from 12 for HOG up to 1208 for a combination of all traditional features. For example, a four-layer network with 20, 13, 17, and 10 neurons on each layer was used with HOG for 12 features, whereas three layers with 200, 100, and 50 neurons were used for LBP with 383 features. Networks with a single layer with ten neurons were utilized for pretrained feature extraction, and transfer learning feature extraction methods with 1000 features.

The testing, validation and training subdatasets were chosen using the same approach as for the training of PSDNet. For the random testing dataset, the 9600 images were divided randomly between training (70 % of images), validation (15 %) and testing (15 %) subdatasets. The soil removal test used the same subdataset proportions as with PSDNet. One material out of 15 was removed from the training and validation subdatasets and used as the testing subdataset (640 images). The remaining images were randomly assigned to the testing (8320 images) and validation (640 images) subdatasets.

## Results

Results for each method are presented in this section. The RMSE on the percentages passing for each sieve is used as the main performance metric.

### PSDNET

PSDNet was assessed using grayscale and color images with sizes of 32, 64, 96, 128, and 160 pixels. Based on **Figure 3**, the best results were achieved for the color images with a size of 160 pixels with an RMSE value of 3.7 %. For the grayscale dataset, the best results were obtained with 128-pixel images with an RMSE of 5.5 %. For a given image size, color images led to better results with RMSE of 3.7 to 7.3 %, compared with RMSE of 5.5 to 8.4 % for grayscale images. On the other hand, for a given image size, the grayscale images required less computing resources for the ConvNet training.

**Figure 3** Results of running PSDNet on grayscale and color images of different sizes.

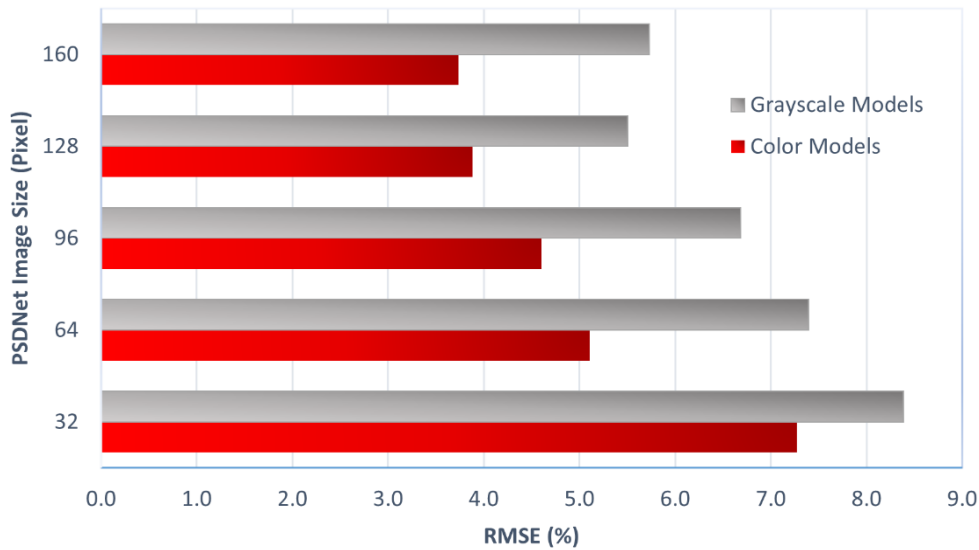

The results for the material removal test for grayscale and color images with PSDNet are presented in figure 4. The best results were obtained for material 4 (passing Bomix and 2C passing) with mean RMSE of 2.8 and 3.9 % for grayscale and color images, respectively. It should be noticed that the best results were obtained for fine-grained materials with similar PSD (materials 4, 6, 7, and 15, figure 1). Several explanations can be suggested for these results, such as the presence of a larger number of similar PSD in the training dataset and less segregation in the fine-grained materials during sample preparation for the photographs. Better performances for finer sieves have also been noted before (e.g. Manashti et al. 2021). The worst RMSE were obtained for material 9 (till) with an RMSE value of 18.5 % for the color dataset. Materials 9, 10 and 11 had the highest RMSE values. Materials 9 and 11 correspond respectively to the finest and coarsest materials. Material 10 has a relatively wide PSD compared to the other materials. This tend to support the hypothesis that better results are obtained with the material removal test when the training dataset includes different materials with similar PSD.

**Figure 4** Results for the material removal test of color and grayscale PSDNet for different image sizes.

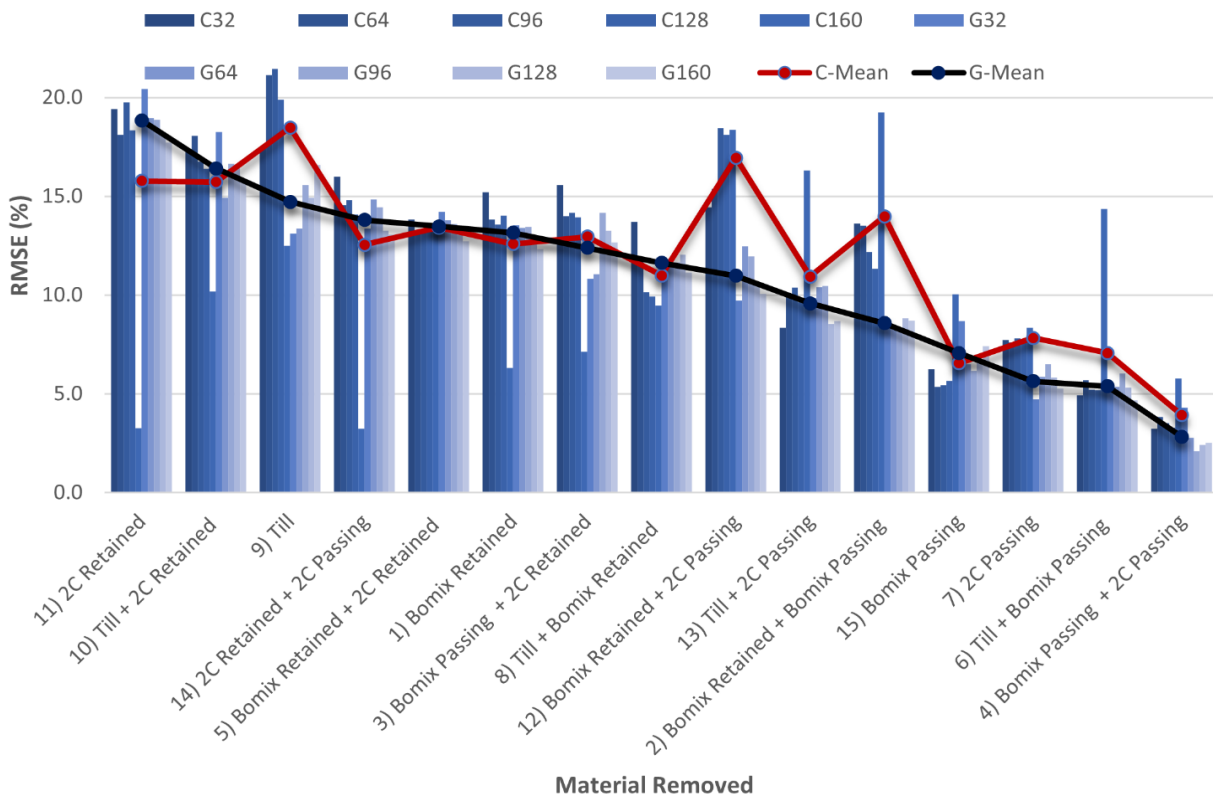

## PSDNET USED AS FEATURE EXTRACTOR

Combining extracted features for various image sizes can improve the prediction accuracy. For this reason, features extracted from the last fully connected layer of the PSDNet networks trained with different images sizes were used as ANN inputs. In this study, 100 features were extracted from the last fully connected layer of PSDNet for both color and grayscale images of 32, 64, 96, 128, and 160 pixels, for a total of 1000 available features. Figure 5 shows the results of feeding different feature combinations to the ANN for color and grayscale images. Using a combination of extracted features for grayscale images can predict the PSD with an RMSE of 3.2 %. Without feature extraction, PSDNet had RMSE values between 5.5 and 8.4 % for grayscale images. For color images, a combination of all PSDNet features led to an RMSE of 2.3 %. In comparison, the RMSE for PSDNet was between 3.7 and 7.3 %. Combining all PSDNet grayscale and color features (1000 features) led to the best results, with an RMSE value of 1.8 %.

The combination of PSDNet features led to good results in the material removal test. As shown in figure 6, combining all PSDNet gray and color features (1000 features) led to mean RMSE values of 3.3 and 18.4 % for material 4 and 11, the best and worst results, respectively. The mean RMSE for the combination of gray, color and both gray and color PSDNet features were 12.2, 10.6, and 9.1 %, respectively. Figure 7 presents the real and predicted PSD for the material removal test with randomly selected images for each of the 15 materials. The subfigure numbers correspond to the PSD in figure 1. Predictions were made by combining the features extracted from PSDNet for all image sizes (32 to 160 pixels), for gray and color images, for a total of 1000 features. Material 12 had the lowest error with an RMSE of 2.7 %. Conversely, in the worst case, the PSD of material 1 was predicted with an RMSE value of 22.1 %. It should be noted that the error for material 1 corresponds approximately to the mean RMSE for the materials with the worst performances in the material removal test for PSDNet (figs 4 and 6).

**Figure 5** Results of combining extracted features of different color and grayscale PSDNet.

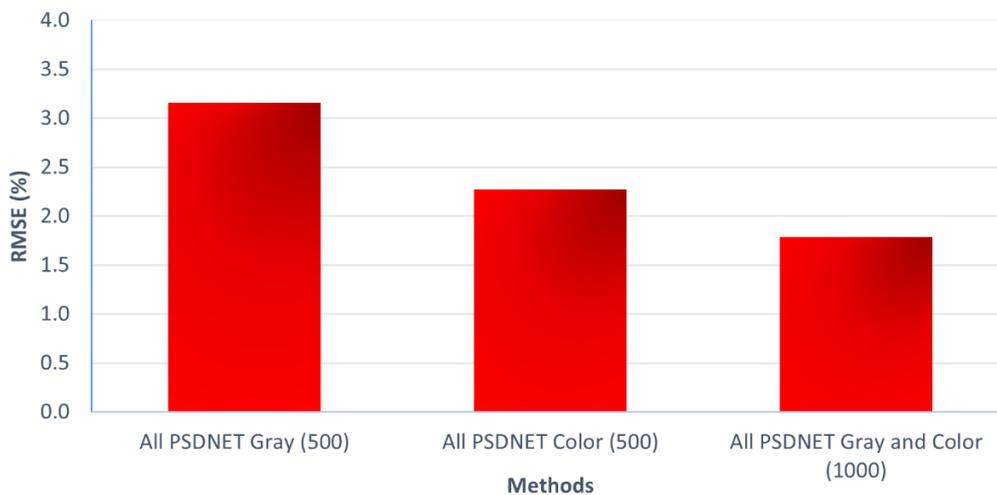

**Figure 6** Results for combination of gray, color, and both gray and color PSDNet extracted features to predict the PSD in the material removal test.

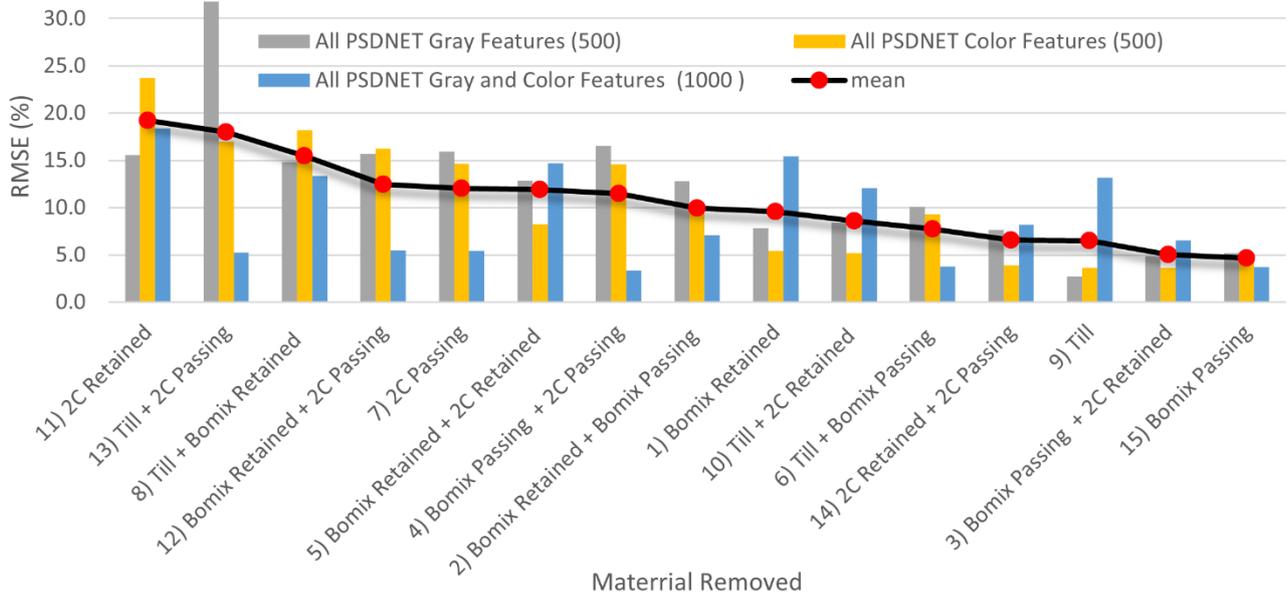

**Figure 7** PSD comparison for randomly selected images for the material removal test. The blue curves were obtained by sieving and the red curves correspond to the network predictions obtained with PSDNet used as a feature extractor for all image sizes and for both grayscale and color images.

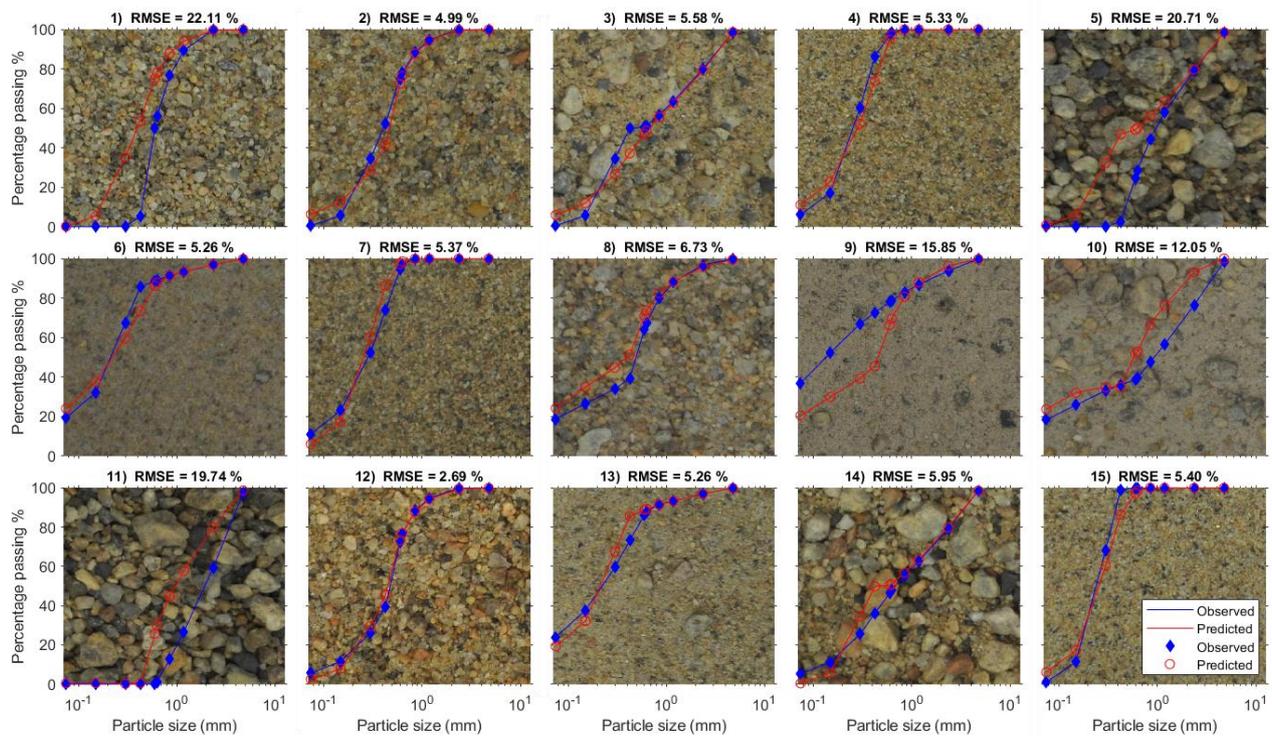

**PRETRAINED MODELS**

Figure 8 compares the RMSE on the percentages passing for the 15 pretrained models. The three methods that were previously introduced are compared (transfer learning, feature extraction of transfer learning, feature extraction). The best results were obtained with transfer learning followed by feature extraction. The best results were achieved with InceptionResNetV2 (RMSE of 1.7 %) followed by NasNetLarge and ResNet101 (RMSE of 2.4 %). GoogLeNet, with an RMSE value of 5.6 %, had the worst performances for feature extraction of transfer learning. Transfer learning of pretrained models led to RMSE values between 4.3% for NasNetLarge and 9.3 % with AlexNet. Feature extraction of pretrained ConvNets achieved RMSE values from 4.0 % with ResNet101 to 6.2 % with GoogLeNet and AlexNet.

**Figure 8** Evaluating of pretrained ConvNet in three-position, including transfer learning, feature extraction, and feature extraction of transfer learning for PSD prediction.

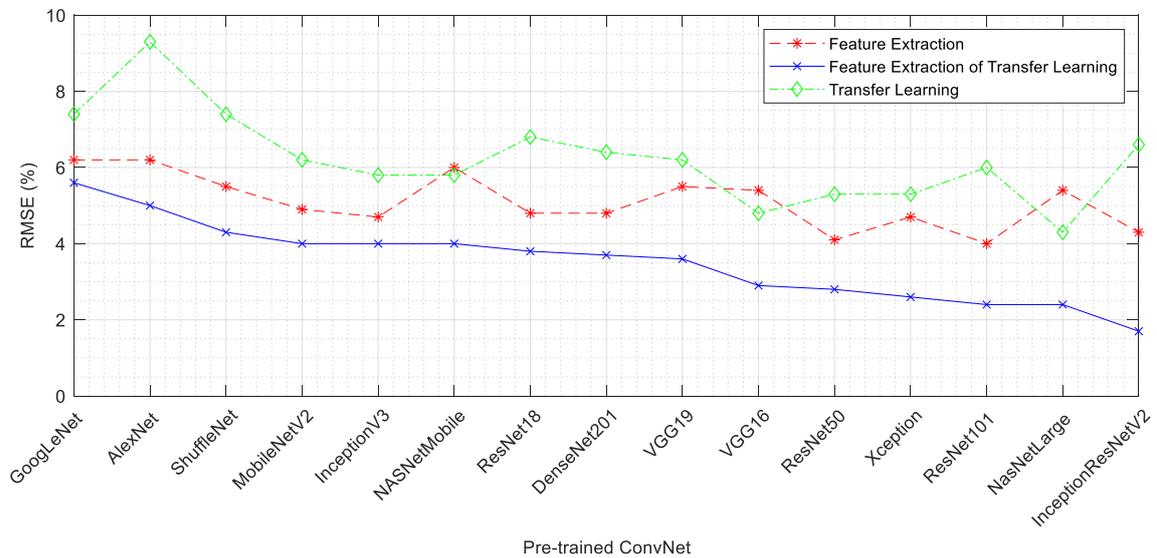

Results for the material removal test for transfer learning models are shown in figure 9. As expected, the RMSE increases in contrast with the random testing dataset. Material 4 (Passing Bomix and 2C Passing) led to the best performances with a mean RMSE for all pretrained networks of 6.1 %, whereas material 11 (2C Retained) led to the poorest performances with an RMSE of 17.0 %. With regards to the performances of each network, Xception and ResNet101, with mean RMSE of 9.3 and 17.3 %, had the best and worst performances, respectively. GoogLeNet could predict the percentages passing for material 4 with a minimum RMSE of 1.5 %, while ResNet101 reached an RMSE of 30.2 % for material 11.

**Figure 9** Results for pretrained models with the transfer learning method for the material-removal test.

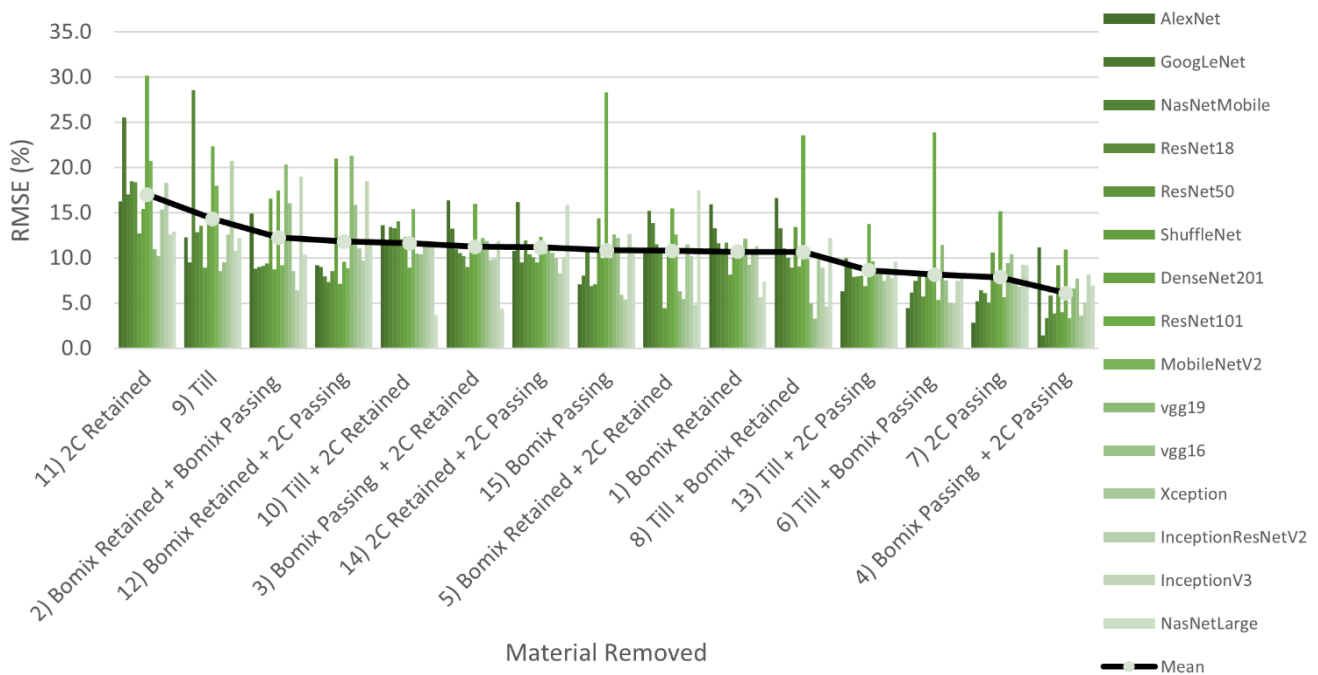

## TRADITIONAL FEATURE EXTRACTION

Results for traditional feature extractions are presented in figure 10. A combination of all methods with 1208 features reached the best results with an RMSE on the percentages passing of 2.7%, followed by entropy, CLBP, and LBP with RMSE of 3.1, 3.3, and 4.0 %, respectively.

As with other techniques, the RMSE increases for the material removal test (figure 11). The best mean RMSE was obtained for material 4.3 %. Material 10 led to the poorest performances with an RMSE of 17.9 %. With respect to the methods, wavelet and a combination of all methods were the top models with a mean RMSE of 3.2 % for material 4. Entropy led to the highest prediction error with a mean RMSE value of 24.7 % for material 10.

**Figure 10** RMSE on percentages passing for traditional feature extraction methods (the number in parentheses following the method name corresponds to the number of features).

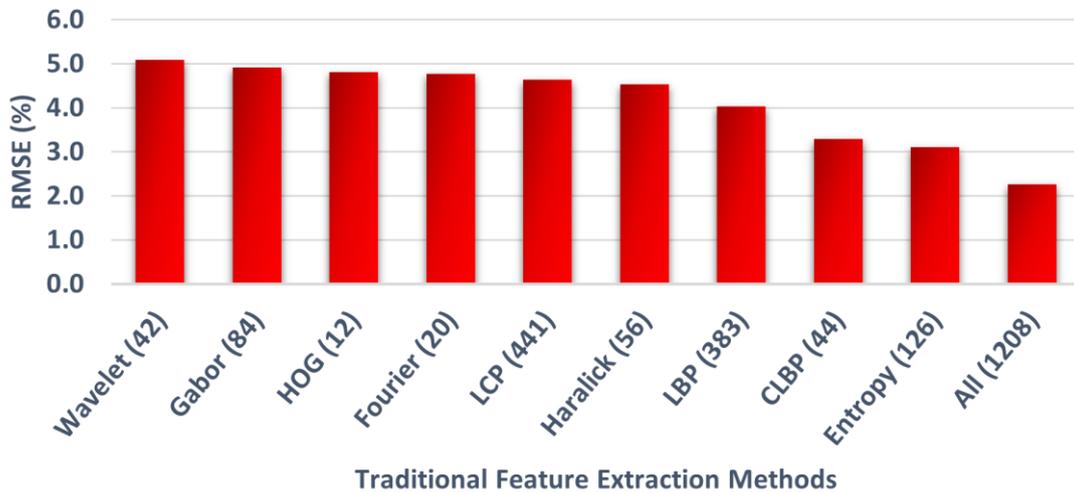

**Figure 11** RMSE for the material removal test with traditional feature extraction methods.

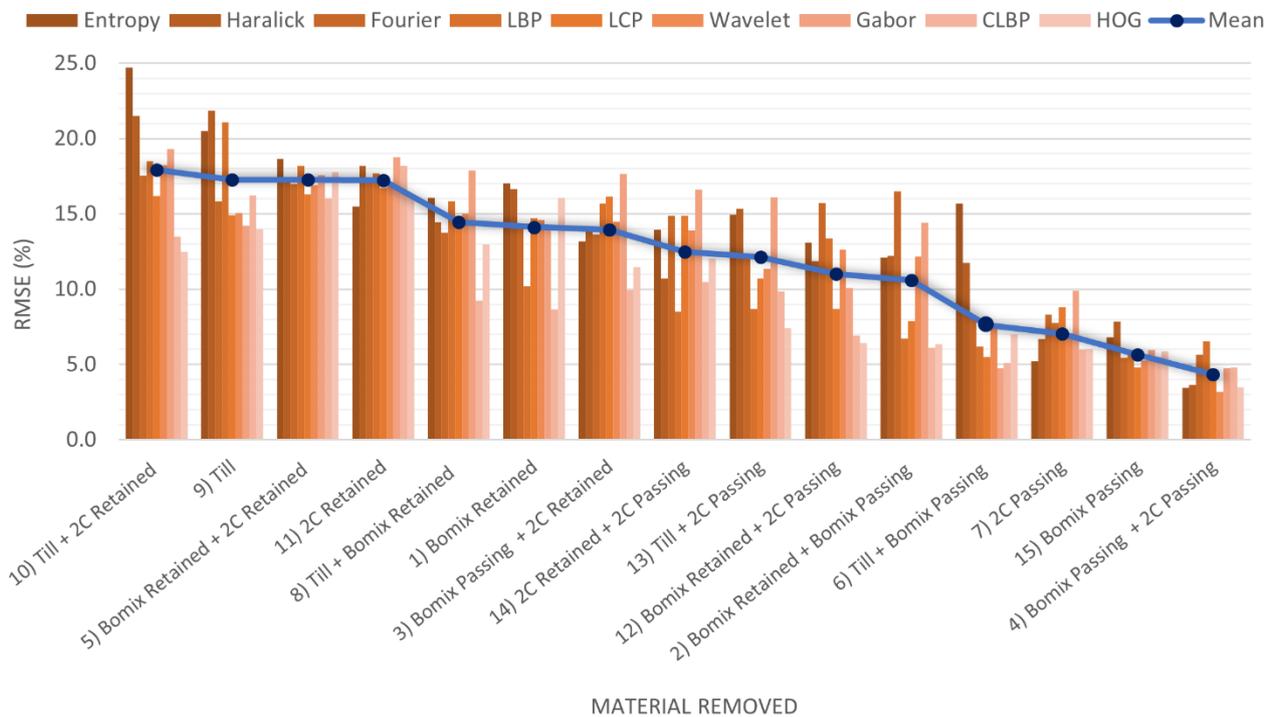

*D$_{50}$*

Error percentages for the $D_{50}$ predicted with PSDNet for random testing of color and grayscale images are shown in figure 12. The error generally decreases with the images size. For the color dataset, the best performances were obtained with the 160-pixel images with a percentage error on $D_{50}$ of 7.9 %. For grayscale images, the lowest error on $D_{50}$ was obtained with the 128 pixel images with an error of 11.2 %

**Figure 12** Percentage error on $D_{50}$ with PSDNet for different image sizes and colors

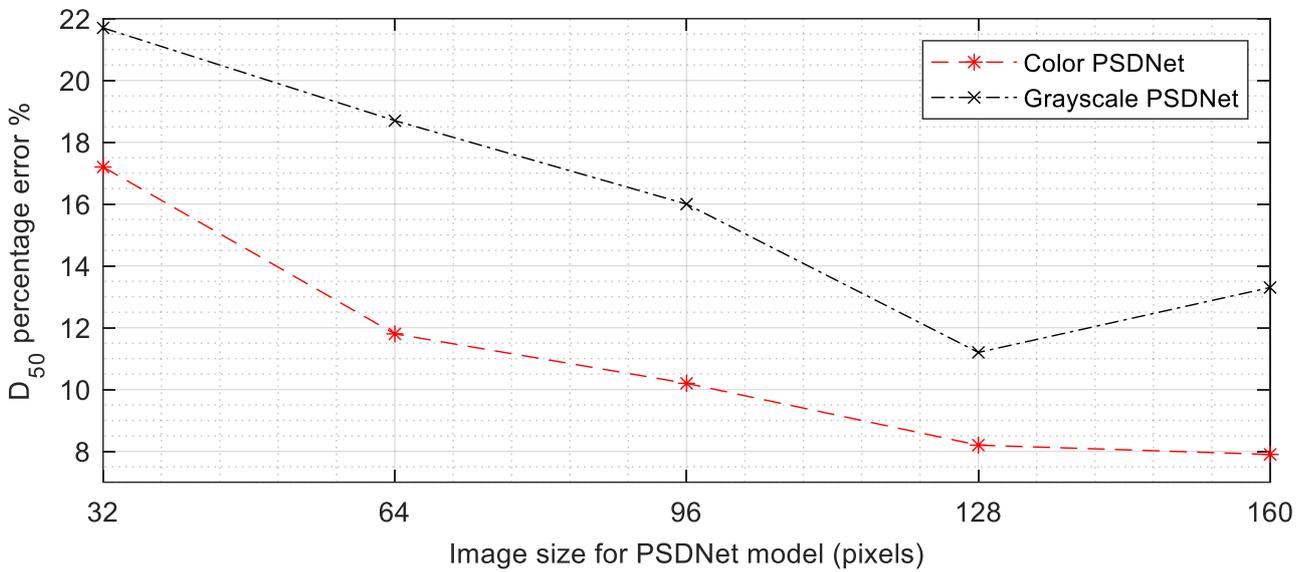

Percentage error on $D_{50}$ for different pretrained ConvNets used with the transfer learning method are shown in figure 13. VGG16, with an error of 11.3%, followed by NasNetLarge and ResNet50 by an error of 12.8 %, reached the best performances. AlexNet reached the worst performances with a percentage error on $D_{50}$ of 26.0 %.

Combining all 1208 traditional features led to a $D_{50}$ percentage error of 4.4 %. Features extracted from PSDNet for all color and grayscale images (1000 features) led to a $D_{50}$ percentage error of 3.4 %. A slightly higher errors were obtained with features for color images (500 features, 4.8 %) and gray scale images (500 features, 4.9 %).

**Figure 13** Relative error on $D_{50}$ for transfer learning with pretrained models.

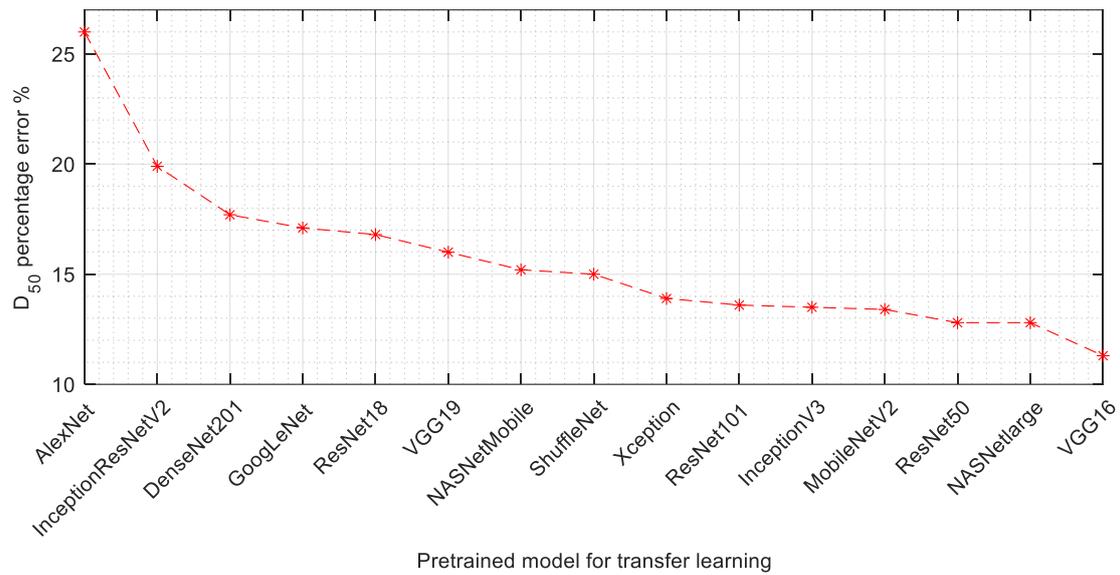

**Figure 14** Prediction error for different sieve sizes for PSDNet with color images of 160 pixels.

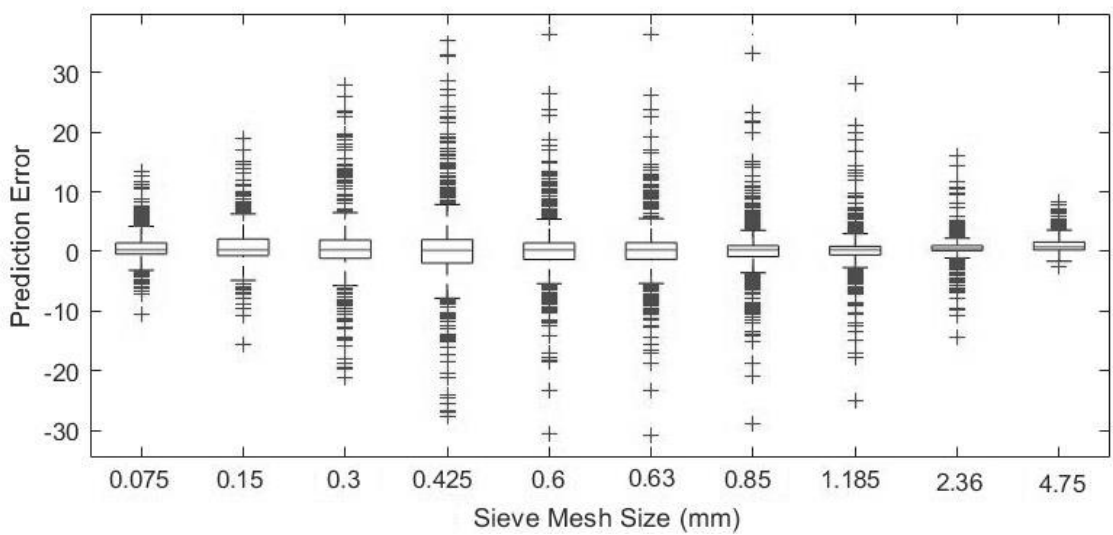

**PREDICTION ERROR CHART**

Figure 14 shows the prediction error for PSDNet in terms of percentages passing for the different sieve sizes for color images of 160 pixel. The prediction error is determined based on the real percentage passing minus the predicted percentage passing for each mesh size. The absolute value of the error is lower for the coarsest sieve (4.75 mm) because the range of percentages passing for this sieve in the dataset is narrow. The largest range of error was obtained for the 0.6 and 0.63 mm sieves. The prediction errors for these sieves range from -31 to 36 %.

Discussion

The mean RMSE on the percentages passing obtained in this study for real photographs of granular materials can be compared to the results presented by Manashti et al. (2021; 2022) for synthetic images. In this study, PSDNet yielded RMSE on percentages passing ranging from 3.7 to 8.4 % for real photographs. The RMSE ranged from 1.7 to 9.3 % for pretrained models, and from 2.3 to 5.1 % for traditional feature extraction. In comparison, for synthetic images of the top surface of particle assemblages in a container, Manashti et al. (2022) obtained RMSE values on the percentages passing ranging between 4.2 and 7.2 % for PSDNet. For pretrained models, the RMSE ranged from 4.6 to 7.5 %. For traditional feature extraction with the same dataset, Manashti et al. (2021) obtained RMSE values ranging between 3.4 and 6.9 %. Therefore, the results obtained for both types of images are generally comparable. Similar conclusion could be drawn from a comparison of the $D_{50}$ errors. It should be noted that the range of PSD that were considered in this study and for the synthetic images used by Manashti et al. (2021; 2022) are different. Manashti et al. (2021; 2022) considered a smaller range of particle size (from 0.075 mm to 1.18 mm, versus 0.075 mm to 4.75 mm in this study), but a much larger number of distinct PSD (53003 versus 15 in this study). A comparison was conducted for very similar range of PSD distributions using local entropy features as ANN inputs by Duhaime et al. (2021). Very similar RMSE were obtained for the real photographs and the synthetic images with RMSE of 3.8 and 3.4 %, respectively.

Using synthetic images for network training can have several advantages. First, it is much easier to generate large datasets of synthetic images than to obtain real photographs and their matching ground truth. For PSD determination, gathering real photographs of granular materials with their PSD is time consuming. Synthetic images can also be used to test the influence on network performances of different dataset parameters (e.g., lighting, camera position). A good example is the comparison of PSDNet performances by Manashti et al. (2022) for synthetic datasets including both views from the top and bottom of a transparent container with performances for datasets including only views from the top. Lower RMSE on the percentages passing can be obtained when both viewpoints are considered (Manashti et al. 2022). Duhaime et al. (2021) have also shown that real and synthetic images can be combined in the same dataset for network training with RMSE that are comparable to those obtained with only synthetic or real images. For other applications, it has been demonstrated that combining real and synthetic images for the training of ConvNets or ANN can increase the generalization potential of networks and avoid overfitting (Nikolenko 2021).

The RMSE values that were obtained in this study can also be compared with published results for real photographs analyzed with other methods, such as direct methods based on image segmentation. RMSE values from the material removal test are probably the most representative performance metric for real applications where the tested material will always be an unseen material. The mean RMSE values that were obtained for the material removal test in this study varied between 2.8 and 19.2 % depending on the method and the material. With direct methods and real photographs, Liu and Tran (1996), and Sudhakar, Adhikari, and Gupta (2006) reached RMSE values on percentages passing ranging between 13 and 36 %, and 15 and 20 %, respectively. In comparison, better performances were obtained in this paper for the best methods with materials with PSD at the center of the grading range in figure 1 (e.g. material 7). The performances for materials with extreme PSD (e.g. material 11) led to performances that are comparable with previously publish results.

Our results suggest that some methods are more computationally efficient than others. The RMSE values obtained for each method are similar, but the computational resources that are required are very different. Traditional feature extraction and pretrained networks used as feature extractors proved to be the most efficient methods. The difference in computational resources is due to the much smaller ANN networks required with feature extraction methods compared to ConvNets, and to the much smaller input size of feature vectors compared to images.

Conclusion

The primary purpose of this paper was to assess the performances of PSDNet to predict the PSD for 9600 photographs of 15 different granular materials. Grayscale and color images of different sizes were tested to predict the percentages passing for 10 sieve sizes. Several others methods were used to predict the PSD, including traditional feature extraction, pretrained ConvNet with transfer learning, feature extraction, and feature extraction of transfer learning. The majority of these methods had been evaluated previously with synthetic datasets by Manashti et al. (2021; 2022).

Each method could predict the PSD accurately. PSDNet could predict the percentages passing with RMSE values of 3.7 and 5.5 % for color and grayscale images, respectively. Feature extraction of transfer learning for the pretrained network InceptionResNetV2 and a combination of all PSDNet gray and color features led to the best results with RMSE values of 1.7 and 1.8 %, respectively.

A material removal test was used to determine the performances of each method when confronted with an unseen material for real applications. All methods could predict the PSD of unseen materials with RMSE values on the percentages passing ranging between 2.1 and 31.8 %. Features extracted from PSDNET for grayscale and color images (1000 features) led to the lowest RMSE (9.1 %), while local entropy led to the highest RMSE (14.1 %). All methods that were presented in this paper were able to predict the PSD of unseen material with performances that were similar, or in some case better, than previously published results obtained with classical methods like segmentation.

Acknowledgments

The authors gratefully acknowledge the funding of Hydro-Québec and the National Sciences and Engineering Research Council of Canada (NSERC) for this project.